# Functional Analytics for Document Ordering for Curriculum Development and Comprehension

Arturo N. Villanueva, Jr.[1] and Steven J. Simske[2]

**Abstract**

We propose multiple techniques for automatic document order generation for (1) curriculum development and for (2) creation of optimal reading order for use in learning, training, and other content-sequencing applications. Such techniques could potentially be used to improve comprehension, identify areas that need expounding, generate curricula, and improve search engine results. We advance two main techniques: The first uses document similarities through various methods. The second uses entropy against the backdrop of topics generated through Latent Dirichlet Allocation (LDA). In addition, we try the same methods on the summarized documents and compare them against the results obtained using the complete documents. Our results showed that while the document orders for our control document sets (biographies, novels, and Wikipedia articles) could not be predicted using our methods, our test documents (textbooks, courses, journal papers, dissertations) provided more reliability. We also demonstrated that summarized documents were good stand-ins for the complete documents for the purposes of ordering.

*Corresponding author* [1]*Email: mrartv@gmail.com  Address: Department of Systems Engineering, Colorado State University, 6029 Campus Delivery, Fort Collins, CO 80523 USA. ORCID: https://orcid.org/0000-0001-5224-9703*
[2] *Email:steve.simske@colostate.edu Address: Address: Department of Systems Engineering, Colorado State University, 6029 Campus Delivery, Fort Collins, CO 80523 USA. ORCID: https://orcid.org/0000-0002-6937-1956*

## 1. Introduction

This research provides, through our investigated algorithms, various reading orders for (1) curriculum development and (2) optimal learning (Villanueva, 2023).

Curriculum development involves sequencing documents such that the transition between topics covered is smooth and builds upon previous material. Optimal learning, on the other hand, aims to maximize functional measurements such as proficiency testing.

For the former, ordering involves arranging topics to ensure that each new concept or skill builds on what is already learned. For example, when developing a language curriculum such as Portuguese, the curriculum may start with basic vocabulary and simple sentences. This foundation allows for the introduction of complex grammar and verb conjugations, setting the stage for advanced topics like idiomatic expressions and nuanced language use. This progression ensures effective language understanding and usage. This is thought to help students grasp new concepts more easily, as they can relate them to a solid foundation of prior knowledge (Richards, 2021). It also allows for the identification of key prerequisites for each subject, ensuring that students are adequately prepared for each new piece of learning.

For the latter, the overarching goal is to increase knowledge by organizing learning materials in a way that is consistent with the principles of cognitive science (Margunayasa et al., 2019; Nurnberger-Haag & Thompson, 2023) and may differ for each student, taking into account his needs and preferences. For example, some students may benefit from the introduction of basic topics that are then coalesced into a broader, more inclusive general topic, similar to the build-up as described for curriculum development. But some students may prefer the opposite approach in which the general topic is covered first, with more detailed concepts covered later as desired. For example, a general topic on microgrids could be followed up by detailed treatises on the duck curve, solar panels, and flywheel energy storage (Arani et al., 2017).

This paper takes into consideration some existing contributions such as those described in the Related Work section below with the goal of providing multiple ways of ordering collections of documents for a variety of purposes, including those described above. In doing so, we also attempt to characterize existing sequences of certain collections such as those of dissertations, textbooks, journals, and courses.

## 2. Related Work

The problem of optimal document ordering is not new. Yet, limited works exist in this area. These include the following which provided some limited guidance to the direction of our study.

Menai, et al. (2018), refer to curriculum ordering as the curriculum sequencing (CS) problem and tackle finding an optimal personalized path for a particular learner using swarm intelligence, and in particular, an ant colony system (ACS) as well as variants of a new method called SwarmRW. Earlier work by Al-Muhaideb & Menai (2011) surveyed existing work in evolution computation (EC) approaches for more personalized curriculum ordering. Such is the tack we adapt in our approach, being aware that not all learners learn the same way.

Research by Ramya Thinniyam (Thinniyam, 2014) tackled document chronology in her doctoral dissertation, "On Statistical Sequencing of Document Collections," focusing on automated statistical techniques for determining the chronological order of corpora using only the words contained in them. With the presumption that documents composed closer in time will be more comparable in their substance, she proposes (1) calculating the distance between document pairs using only their word features, and (2) estimating the optimal document order based on these distances. The research examined different types of distances that can be determined between document pairs and introduced methods for sequencing a set of documents based on their pairwise distances. While ordering in this context focuses on the chronology of documents being written, the concept of using contextual embeddings has been applied to our research.

Koutrika, et al., in "Generating Reading Orders over Document Collections" (Koutrika et al., 2015), introduce a system that autonomously arranges a set of documents into a hierarchical structure, progressing from the more general to the more specific content. The system provides users with the flexibility to select a preferred sequence for reading the documents. This approach to content consumption deviates from conventional ranked lists of documents which rely on their relevance to user queries, and static interfaces. Using topics, the authors propose a collection of algorithms that address this problem and assess their performance alongside the generated reading trees. The Koutrika approach uses topics generated via LDA, the same approach we have adopted in this study.

More recently, in Forestier et al. (2022), the authors posit the translation of human "intrinsically motivated spontaneous exploration" learning approaches into machine learning algorithms. The authors term the algorithmic approach Intrinsically Motivated Goal Exploration Processes (IMGEP) with the intent of empowering similar properties of autonomous learning in machines. The team demonstrated their results in automatically generated learning curricula that allowed "the discovery of diverse skills that act as stepping stones for learning more complex skills."

## 3. Research Goals

Informational material is designed to provide knowledge on a particular subject and can be presented in various forms, such as lectures, training modules, or book chapters. Depending on the purpose, readers may choose to consume this material in different orders or subsets. For example, someone who is already familiar with a particular topic may choose to read only specific chapters or sections of a book or training module, rather than starting from the beginning.

However, the default consumption order for informational material is usually in the order of chapter or lecture number. This order is often chosen by the author, editor, or publisher to provide a logical progression of ideas, and to ensure that readers have the necessary background knowledge before moving on to more advanced concepts.

That said, not all orders are equal, and some may be better suited to certain purposes. For example, if the goal is to gain a general overview of a subject, reading the introduction and conclusion chapters of a book or training module may be sufficient. On the other hand, if the goal is to develop a deep understanding of a subject, reading all the chapters in a logical sequence may be necessary.

In some cases, orders may not be predetermined, such as when assembling a curriculum from existing documents. This can involve selecting the most relevant chapters or sections from multiple sources and organizing them in a logical sequence.

The goal of our research is to develop ordering algorithms that improve comprehension and facilitate the creation of effective training curricula. By understanding the most effective order in which to present information, educators can improve student learning outcomes and ensure that key concepts are fully understood.

Additionally, predicting (or suggesting) the order of chapters in a textbook, lecture, or other set of documents may help readers to better plan their study schedule and provide a framework for understanding how the different pieces of information fit together.

This paper takes into consideration existing contributions such as those described in the Related Work section above with the goal of providing multiple ways of ordering collections of documents for a variety of purposes, including those described in the Introduction. In doing so, we also characterize existing sequences of certain collections such as those of dissertations, textbooks, journals, and courses.

## 4. Process / Tasks

While predicting the order of chapters as they were composed by the author may be an interesting exercise and is an offshoot of our research, the main intended use is for generating a proposed reading order.

### 4.1. Overview

Figure 1 summarizes the many combinations of the algorithms we used to investigate their effectiveness. Starting from the Document Set input from the upper left, the workflow splits into three main branches: (1) On the left side, the documents are used to generate comparison matrices used for generating order. (2) On the right side, orders are generated using entropy and Kullback-Leibler divergence (KL divergence) calculations based on the Latent Dirichlet Allocation (LDA) topics. (3) The middle path preprocesses the documents through summarization and feeds those to the first two paths. Various metrics are then collected and compared. Section 4.4 describes the entire process in detail.

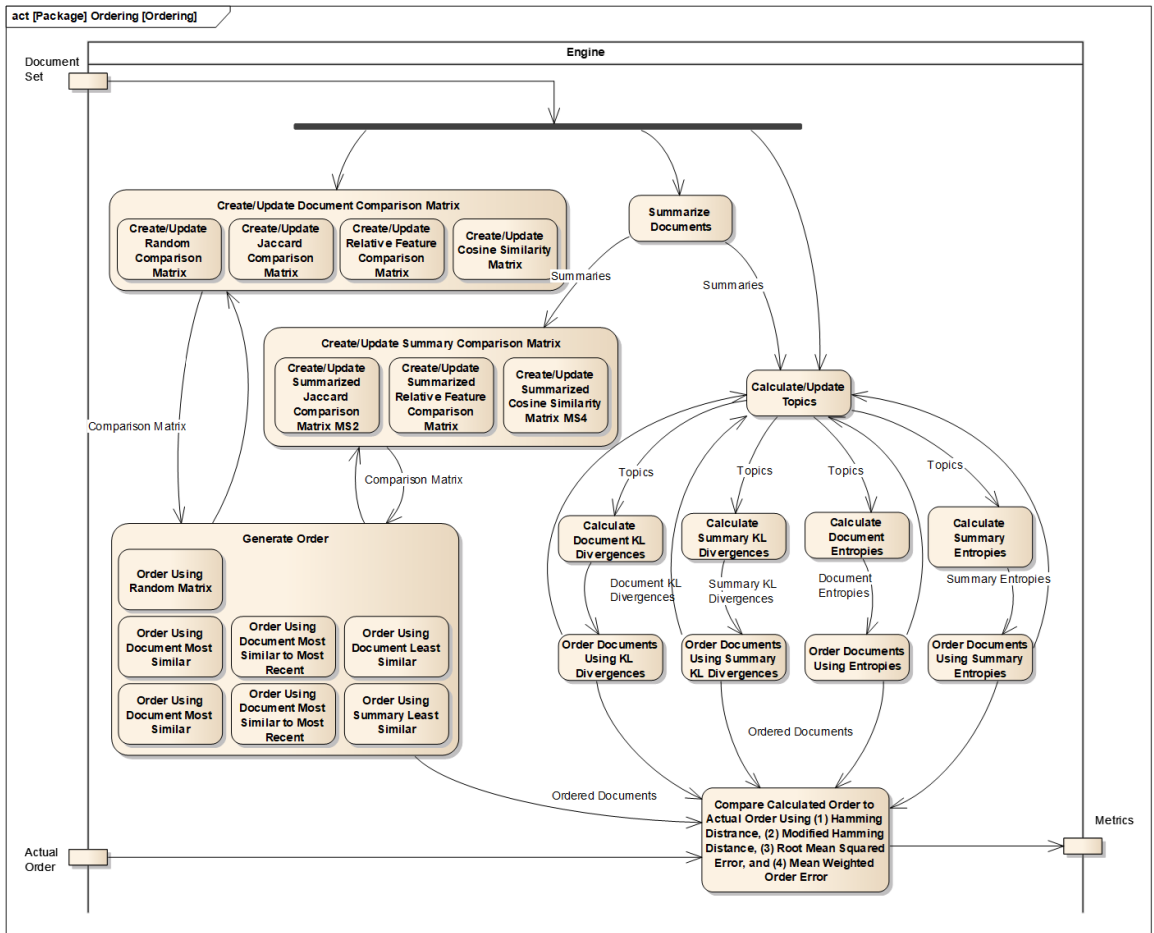

**Figure 1. Overall Ordering Process**

### 4.2. Data Set

To have a good understanding of how our algorithms perform, it is necessary to employ several data sets and types of data sets for testing. The test corpora we used consists of courses, dissertations, journals, and textbooks and control corpora are biographies, novels, and Wikipedia articles. In choosing the test corpora (specifically the textbooks and possibly more so courses and dissertations), we expected that in general, we could detect an order in the documents. Those that do not appear to require or have a specific order, we would attempt to explain away.

In choosing these test corpora, we have endeavored to limit the minimum number of chapters to six, giving us at least $6! = 720$ sequence variations, reducing perfect ordering simply by randomly guessing. (e.g., 3 would be too easy to guess with only 6 different sequence variations.)

**Courses**. While course lectures are often meant to be progressive and order-dependent (i.e., lectures built upon previous lectures), skipping lectures, depending on the course, may or may not significantly affect comprehension. Most of the courses we analyzed were donated by Prof. Simske of the Department of Systems Engineering.

**Dissertations**. Dissertations are highly appropriate for ordering exercises because usually, there is a single focused topic and each chapter of a 100-300-page doctoral dissertation is sufficiently long for analysis. The dissertations we analyzed were retrieved from the Colorado State University repository for doctoral dissertations and master's theses, though we limited our selection to doctoral dissertations as they tend to be longer and much richer in content.

**Journals**. Journals are periodicals that accept authors from various organizations that do not necessarily work on research related to each other. Papers published in journals are therefore self-contained and do not rely on other papers within that publication. Related subject areas may be grouped within chapters while no real required order of reading is suggested, either explicitly or implicitly. However, we surmise that papers have a loose order in which more general topics are covered earlier and more specific ones are covered later.

**Textbooks**. While textbooks can and are often presented in the order in which they are meant to be read, instructors using those textbooks for instruction often jump from chapter to chapter in an alternative order, sometimes skipping chapters altogether. This indicates that chapters can and do often stand on their own. The collection of textbooks we used were obtained from the Colorado State University library, donated by professors, or sourced from the public domain.

In addition to the above, we use several control documents to provide contrast.

**Biographies**. While biographies rely on the progression of a story and thus the reading a middle chapter before reading the introduction typically won't make for a sensible narrative, their order is time-based and not information-based. That is, in general, the prerequisite for understanding a later chapter relies on the understanding of previous chapter for the reason that something happened earlier in time that caused something later in time. Foundational knowledge presented in earlier chapters rely on time but do not necessarily add depth to the whole. Our algorithms, on the other hand, are depth-based in that entropy of words and similarity of word structures are the pieces examined for relationships. All biographies were retrieved from Project Gutenberg (*Project Gutenberg*, n.d.).

**Novels**. While fiction in nature, novels are time-based narratives as with biographies, and thus have similar limitations. Novels often have characters that appear periodically, may have several interwoven sub-narratives, each of which might be sequential but not necessarily. All novels were retrieved from Project Gutenberg (*Project Gutenberg*, n.d.).

**Wikipedia Articles**. Wikipedia articles are, by design, meant to stand independent of one another even if hyperlinking to other articles is encouraged. With millions of authors, this is also the de facto structure. In principle, however, it is impossible for all articles to present all requisite knowledge within the articles themselves. For example, the article on Petri Nets makes an assumption that the reader is able to understand set and graph theory and able to read mathematical notation pertinent to them.

### 4.3. Data Preparation

Each of the data sets described above (genres) were treated as collections of documents that were split into logical documents. Dissertations, textbooks, novels, and biographies were split into the author-identified chapters. Dissertations contained a minimum and usually not more than six chapters each and textbooks, novels, and biographies had a much wider range of numbers of chapters. Course sets were split into weekly lectures and depending on whether they occurred during the summer or a regular semester, were thirteen or fifteen weekly lectures long. Wikipedia article sets were split into individual articles. And journals were split into their component articles.

Preparatory processing steps were first performed including removal of stop words (using the SciKit-learn library (Pedregosa et al., 2011) plus ['abstract', 'introduction', 'chapter', 'figure', 'fig', 'table']) and lemmatization of each document. We then created a document-term frequency matrix from the remaining terms.

### 4.4. Sequencing Method

#### 4.4.1. Similarity Matrix-based Content Sequencing Method Using Complete Documents

**Step One: Generate Similarity Matrix**

Given an array of $n$ documents $D$, we define its similarity matrix $\mathbf{S}_D = (\gamma_{ij})$ which is an $n \times n$ symmetric matrix. If each $\gamma_{ij}$ represents how the row document $d_i$ compares with the column document $d_j$ (or vice versa), we have

$$\gamma_{ij} = \begin{cases} 1 & \text{if } i = j \\ c_{ij} & \text{if } i \neq j \end{cases} \tag{1}$$

All the values $c_{ij}$ depend on how comparisons are handled. We use three methods for comparison with varying results: Cosine Similarity (*Cosine Similarity - an Overview*, n.d.), Jaccard similarity (Niwattanakul et al., 2013), and a metric we call the relative probability similarity.

**Cosine Similarity.** Cosine similarity is a common method of comparing two documents based on their content. The Cosine Similarity of two documents $a$ and $b$ is calculated:

$$s_C(a,b) = \cos(a,b) = \frac{a \cdot b}{\|a\|\|b\|} = \frac{\sum_{i=1}^n a_i b_i}{\sqrt{\sum_{i=1}^n a_i^2}\sqrt{\sum_{i=1}^n b_i^2}} \tag{2}$$

For our example, the similarity of the chapters using Cosine Similarity may look like Equation (3).

$$s_C(a,b) = \begin{bmatrix} 1 & s_C(1,2) & \ldots & s_C(1,n) \\ s_C(1,2) & \ldots & \ldots & s_C(2,n) \\ \ldots & \ldots & \ldots & \ldots \\ s_C(1,n) & s_C(2,n) & \ldots & 1 \end{bmatrix} \tag{3}$$

**Jaccard Similarity.** Jaccard similarity is calculated by using the overlap of the words that comprise the documents:

$$s_J(a,b) = J(A,B) = \frac{|A \cap B|}{|A \cup B|} \tag{4}$$

where $A$ and $B$ are the sets of words (features) that are contained in the corresponding $a$ and $b$ documents. The matrix using Jaccard similarity thus looks like Equation (5).

$$s_J(a,b) = \begin{bmatrix} 1 & s_J(1,2) & \ldots & s_J(1,n) \\ s_J(1,2) & 1 & \ldots & s_J(2,n) \\ \ldots & \ldots & \ldots & \ldots \\ s_J(1,n) & s_J(2,n) & \ldots & 1 \end{bmatrix} \tag{5}$$

For our calculations, we restricted ourselves to single words and not multi-$n$-grams.

**Relative Probability Similarity.** Our third way of comparing documents is by a feature-wise comparison. We define the similarity of two documents by:

$$s_R(a,b) = 1 - \sum_{i=1}^n |f_i(a) - f_i(b)| \tag{6}$$

where $f_i(d)$ is the $i$th feature (frequency of the word) of the document vector $d$.

We considered but opted against an alternate definition,

$$s_R(a,b) = \sum_{i=1}^n \alpha \beta \tag{7}$$

where

$$\alpha = \begin{cases} 0 & \text{if } f_i(a) = 0 \\ \log(f_i(a)) & \text{if } f_i(a) \neq 0 \end{cases} \quad \text{and} \quad \beta = \begin{cases} 0 & \text{if } f_i(b) = 0 \\ \log(f_i(b)) & \text{if } f_i(b) \neq 0 \end{cases}$$

as $s_R$ in this case did not accurately represent similarity when documents were subjectively evaluated.

For our chosen method (6), our document similarity matrix is represented as in Equation (8).

$$s_R(a,b) = \begin{bmatrix} 1 & s_R(1,2) & ... & s_R(1,n) \\ s_R(1,2) & 1 & ... & s_R(2,n) \\ ... & ... & ... & ... \\ s_R(1,n) & s_R(2,n) & ... & 1 \end{bmatrix} \quad (8)$$

**Step Two: Generate the Sequence**

Multiple ways of generating sequences were investigated. In all sequence-generating schemes, it was important to have a baseline comparison sequence. For this purpose, we generated 100 random sequences and calculated the metrics for each of them. The means of these metrics for each of the different sequences and compared with the results gotten for each following three methods.

**Sequence Most Similar to All Selected Documents.** This method of generating a sequence assumes that the first document is the most general and therefore most similar to all the other documents in the set. Ensuing (sequentially following) documents chosen to build the sequence are chosen for their similarity to the already chosen documents, relying on the concept that transitions are smoothly bridged between the documents.

The sequencing of the documents is initially based on the matrix as described in one of the methods in 0 and is iterative. Operating on $S_D$ as defined above and referring to (1), we select document

$$d_{next} = \left\{ d_{max} \middle| \{d_{max} \in D\} \wedge \left\{ d_{max} \cong \min_{d_i}(\bar{\gamma}_{ij}), i \neq j \right\} \right\} \quad (9)$$

as the first element in the computed sequence. $d_{max}$ is removed from $S_D$, which is then recalculated as an $(n-1)x(n-1)$ $S'_D$ matrix with the remaining documents using the chosen Step 1 matrix similarity method.

The next document is chosen similarly from the $S'_D$ matrix with the exception that each of the remaining documents is compared with the mean of the values of the already chosen documents. Iterations are repeated until the original $S_D$ is reduced to a $1x1$ matrix. At which point, a complete sequence will have been generated.

**Sequence Most Similar to Most Recent Document.** This method shares a lot of similarities with **Sequence Most Similar to All Selected Documents**, with a deviation in the way the iterative matrices are calculated. Rather than the second and ensuing documents chosen from the resulting $S'_D$ matrix with each of the remaining documents compared with the mean of the values of the already chosen documents, they are compared to only the last chosen document. As with **Sequence Most Similar to All Selected Documents**, iterations are repeated until the original $S_D$ is reduced to a $1x1$ matrix. While this method of sequencing also picks the most general document as the starting point, ensuing documents are chosen to be most similar to the last previously chosen document only, attempting to create smooth transition from document to document.

**Sequence Least Similar to All Selected Documents.** The third way of sequencing using the similarity matrix method can be thought of as the opposite of the **Sequence Most Similar to All Selected Documents** and attempts to generate order with the idea that the most specific topics are necessary building blocks in order to be able to grasp the information presented by the document that is the culmination of the introductory material.

Again, sequencing the documents is initially based on the matrix as described in one of the methods in 0 and is iterative. Operating on $S_D$, we select document

$$d_{next} = \left\{ d_{min} \middle| \{d_{min} \in D\} \wedge \left\{ d_{min} \cong \min_{d_i}(\bar{\gamma}_{ij}), i \neq j \right\} \right\} \quad (10)$$

as the first element in the computed sequence. $d_{min}$ is removed from $S_D$, which is then recalculated as an $(n-1)x(n-1)$ $S'_D$ matrix with the remaining documents using the chosen Step 1 matrix similarity method.

**Variation: Similarity Matrix-based Content Sequencing Method Using Summaries**

While the above methods were first employed on the full texts of the documents, we were also interested in whether operating on the summaries of these texts would provide us with similar results. A strong positive correlation using reasonably-sized summaries could provide enhanced processing speed at worst. And at best, operating on summaries could provide even better sequence ordering.

We opted for an extractive summarizer to guarantee compact verbatim extractions that used words that existed in the original text. Such was necessary so that 1) topic extraction would operate on identical words to those that exist in the full text and 2) similarity comparisons such as Jaccard and Cosine Similarity could similarly operate on the summaries as with the full documents without the use of expensive word embeddings such as Word2Vec (Radim

Řehůřek, n.d.) or GloVe (Pennington et al., 2014). The Luhn heuristic extractive summarization algorithm (Verma et al., 2019) was chosen from the many algorithms provided in the Sumy library (Python Software Foundation, 2022) based on its performance in experimentation compared with SumBasic, LexRank, LSA, TextRank, Edmundson, and Kullback-Leibler. Luhn is based on TF*IDF (Term Frequency-Inverse Document Frequency) and discounts stop words.

### 4.4.2. Entropy Ordering Method of Complete Documents

Entropy can be used in at least two potential ways as methods for ordering. The first is the idea that introductory corpora are more general and provide more general, superficial treatise of the topics covered by the rest of the corpora. The second is the idea that less entropic corpora are prerequisite material for understanding the most entropic corpora.

**Step One: Generate the Topics**

For topic generation, we used Latent Dirichlet Allocation (LDA), described in the seminal paper by Blei, Ng, and Jordan (Blei et al., 2003) and implemented in the Gensim library (Řehůřek & Sojka, 2010). LDA is highly dependent on the number of topics passed as a parameter, but in order to limit the number of variables for this research, we standardized by choosing the number of topics to be 20% of the total number of documents as a reasonable number (see 5.1.1). For example, an introductory systems engineering textbook may cover the topics of requirements engineering, architecture and design, development and integration, verification and validation, and maintenance. A text on chess may cover history, rules of chess, openings, middle game, and end game. Undoubtedly, this is a very rough approximation and different corpus collections have various topics. Regardless, optimizing for the ideal number of topics for the use of entropy calculations is left as a topic for future research.

We define the topic array as

$$T = \begin{bmatrix} t_1 \\ t_2 \\ \dots \\ t_p \end{bmatrix}, \text{ and each } t_i = \begin{bmatrix} w_1 & w_2 & \dots & w_m \end{bmatrix} \quad (11)$$

where $w_i$ is the $i$th keyword-probability tuple of the $m$ keyword-probability tuples that represent the document. $p$ is the number of topics chosen and $t_i$ is the $i$th topic vector with $m$ word tuples. For illustrative purposes, topics generated by LDA for the 'Simske Functional Applications' book are listed in Table 1. It shows a sample matrix $T$ (extracted from our test data set 'Simske Functional Applications' book). Each column represents a topic defined by a set of tuples of words and their percentage contribution to defining the topic.

**Table 1. Sample topic matrix, T, from corpus 'Simske Functional Applications' book**

| Topic 1 ($t_1$) | Topic 2 ($t_2$) | Topic 3 ($t_3$) |
|---|---|---|
| (cluster, 0.03477) | (text, 0.01935) | (summarization, 0.01954) |
| (class, 0.02063) | (learn, 0.01473) | (weight, 0.01823) |
| (data, 0.01848) | (accuracy, 0.01276) | (sentence, 0.01678) |
| (classification, 0.01524) | (translation, 0.01266) | (word, 0.01666) |
| (equation, 0.01086) | (language, 0.01191) | (text, 0.01565) |
| (distance, 0.01021) | (document, 0.01093) | (document, 0.01518) |
| (train, 0.009958) | (data, 0.009477) | (count, 0.01059) |
| (categorization, 0.009648) | (order, 0.009465) | (example, 0.00941) |
| (example, 0.008913) | (read, 0.009278) | (use, 0.008444) |
| (score, 0.008463) | (example, 0.008173) | (language, 0.008234) |

Based on these definitions, each document (chapter) in the corpus collection is represented best by a topic, referred to as the dominant topic. Table 2 illustrates this.

**Table 2. Sample Document-Topic assignment of corpus 'Simske Functional Applications' book**

| Document | Chapter Title | (Dominant) Topic |
|---|---|---|
| Chapter 1 | Linguistics and NLP | Topic 3 |
| Chapter 2 | Summarization | Topic 3 |
| Chapter 3 | Clustering, Classification, and Categorization | Topic 1 |
| Chapter 4 | Translation | Topic 2 |
| Chapter 5 | Optimization | Topic 2 |
| Chapter 6 | Learning | Topic 2 |
| Chapter 7 | Testing and Configuration | Topic 2 |

**Step Two: Generate the Sequence**

Calculate Corpus Entropy Given Topics: After evaluating all the corpora documents through Gensim to generate the topics $(0..p-1)$, with each topic, the next step was to calculate the makeup of each chapter

$$D_W = \begin{bmatrix} d_{W_1} \\ d_{W_2} \\ ... \\ d_{W_n} \end{bmatrix} \quad (12)$$

where $d_{W_i}$ is the document-term vector for each document $i$ of the collection of $n$ documents.

Using the document-term matrix $D_W$ and Topics table $T$, we generate the document-topic matrix $D_T$, which is comprised of the percentage of the topics that are discussed in every given document in $D$. The Document topic matrix explains how k topics are distributed in the n documents. From our $D$ (Table 1) and $T$ (Table 2), we generate the following document topic matrix $D_T$.

$$D_T = \begin{bmatrix} d_{T_1} \\ d_{T_2} \\ ... \\ d_{T_n} \end{bmatrix} \quad (13)$$

where $d_{T_i}$ is the document vector with $p$ topics as features (where $p$ (= 20% number of documents) is the number of topics assumed for this research as described above) for each document $i$ of the collection of $n$ documents. A sample matrix is shown in Equation (14).

$$D_T = \begin{bmatrix} (d_1, t_1) & ... & (d_1, t_p) \\ ... & ... & ... \\ (d_1, t_1) & ... & (d_n, t_p) \end{bmatrix} \quad (14)$$

**Sequence Most to Least Entropic.** Entropy, defined in (15), measures the 'amount of disorder' of a system, or in this case, a document (*Scipy.Stats.Entropy 1.9.1*, n.d.; Shannon, 1948). Conceptually, ordering a set of documents starting from a most entropic to least entropic relies on the idea that a document is more entropic when more topics are covered. The first documents are considered more entropic because they are more of an introductory nature and therefore cover broader topics. The later documents cover more specific topics.

$$H = -\sum_{i=1}^{N} p(x_i) \log p(x_i) \quad (15)$$

This step of sequencing using entropy relies on the topics generated in 0 and more specifically, the Document-Topic Matrix, $D_T$, and as illustrated in Equation (13 and (14).

$$D_H = \begin{bmatrix} H(X_{T1}) \\ H(X_{T2}) \\ ... \\ H(X_{Tn}) \end{bmatrix} \quad (16)$$

The first document chosen is the most entropic ((18) as measured against the topics that documents have been clustered using LDA. Once this first document has been removed, LDA is re-applied to the remaining documents. The second document is then chosen to be the most entropic of the remaining set. The process is repeated until all the documents from the first set have been consumed to produce the ordered set. In general, we have

$$d_{next} = \left\{ d_{max} \middle| \{d_{max} \in D\} \wedge \left\{ d_{max} \cong \max_{d_i}(H_i) \right\} \right\} \tag{17}$$

Note that the generated ordered set of documents $\{d'_1, d'_2, \ldots, d'_2\}$ does not guarantee that $H_{d'_1} > H_{d'_2} > \cdots > H_{d'_n}$ because in each iteration, the LDA and entropy are recalculated amongst the remaining document peers.

**Sequence Least to Most Entropic.** This sequence varies from the preceding sequence generator only in that the new sequence is generated such that at each iteration, the minimum entropy document is chosen (18). Conceptually, the documents with the least entropy are more specific and therefore may serve as satisfying prerequisite information before those documents with higher entropy.

$$d_{next} = \left\{ d_{min} \middle| \{d_{min} \in D\} \wedge \left\{ d_{min} \cong \min_{d_i}(H_i) \right\} \right\} \tag{18}$$

And similarly, here, we do not have a guarantee that $H_{d'_1} < H_{d'_2} < \cdots < H_{d'_n}$.

**Sequence Most to Least KL-Divergent.** The Kullback-Leibler divergence (KL divergence), also known as relative entropy, is defined as

$$KLD(p \parallel q) = \sum_{i=1}^{N} p(x_i) \log \frac{p(x_i)}{q(x_i)} \tag{19}$$

and correspondingly, for the array of document $D$,

$$D_{KL} = \begin{bmatrix} KLD(X_{T1}) \\ KLD(X_{T2}) \\ \ldots \\ KLD(X_{Tn}) \end{bmatrix} \tag{20}$$

The KL divergence calculates the amount of information lost by approximating one distribution with another (Kurt, 2019, 2017). In our case we approximate with a uniform distribution $q$ with the same number of elements as $p$, corresponding to the number of topics.

We follow the same procedures as described in **Sequence Most to Least Entropic** but using the KLD formula

$$d_{next} = \left\{ d_{max} \middle| \{d_{max} \in D\} \wedge \left\{ d_{max} \cong \max_{d_i}(KLD_i) \right\} \right\} \tag{21}$$

**Sequence Least to Most KL-Divergent.** Finally, we follow the same procedure as described in **Sequence Least to Most Entropic** with the following formula

$$d_{next} = \left\{ d_{min} \middle| \{d_{min} \in D\} \wedge \left\{ d_{min} \cong \min_{d_i}(KLD_i) \right\} \right\} \tag{22}$$

### Variation: Entropy Ordering Method of Summarized Documents

In their paper, "Summarization Assessment Methodology for Multiple Corpora Using Queries and Classification for Functional Evaluation," Wolyn and Simske (Wolyn & Simske, 2022) demonstrated that summaries can be good substitutes for complete texts. We expand on this by experimenting with summaries with our analyses and comparing the results with those gotten from analyzing their complete document counterparts.

### 4.5. Metric Selection

While our generated sequences provide suggestions for ordering of corpora depending on the purpose, where possible (such as with chapters in a textbook, chapters in a dissertation, and lectures in a course), we compare them with the author's or instructor's (or curriculum designer's) intent as the gold standard. However, since 'goodness' of order, even with comparisons to the gold standard, is at least partially subjective, we use a variety of metrics to add some objectivity. For illustration purposes, we define the following example: $A_1 = [1\ 2\ 3\ 4\ 5]$, $B_1 = [2\ 3\ 4\ 5\ 1]$, and $B_2 = [5\ 4\ 1\ 2\ 3]$

**Normalized Hamming Distance (NHD).** The Hamming Distance between two sequences of equal length is the number of positions that differ in value, and thus the minimum number of substitutions needed on either sequence to make them identical to the other (Hamming, 2012):

$$HD(A,B) = \sum c_i \tag{23}$$

where

$$c_i = \begin{cases} 0 \text{ if } a_i = b_i \\ 1 \text{ if } a_i \neq b_i \end{cases}$$

A shortcoming of this metric is it uses pairwise comparisons for each position in the two sequences, resulting in a comprehensive comparison that does not indicate how far apart they are. A small perturbation that shifts the order forward or backward by one position produces inordinate errors. For our example, $HD(A_1, B_1) = 5$, which is the same as $HD(A_1, B_2)$. Clearly, however, we can see that $B_1$ is a better ordering approximation of $A_1$ than $B_2$ is to $A_1$. We nonetheless include this metric for comparison with other metrics. For our purposes, we normalize the Hamming Distances to get a value between [0, 1].

**Normalized Modified Hamming Distance (NMHD).** A Modified Hamming Distance was included in our research measurements to account for the shortcoming of the Hamming Distance. Our implementation takes into consideration the distances between each feature in the two sequences in question (Simske & Vans, 2021):

$$MHD(A,B) = \sum |a_i - b_i| \tag{24}$$

For our example, $MHD(A_1, B_1) = 1 + 1 + 1 + 1 + 4 = 8$ but $MHD(A_1, B_2) = 4 + 2 + 2 + 2 + 2 = 12$, capturing the superiority of $B_1$ over $B_2$. As with the HD, we normalize the Modified Hamming Distance to get a value between [0, 1].

**Normalized Root Mean Square Error (NRMSE).** The third metric we used is the Root Mean Square Error, which is a second order version of the Normalized Hamming Distance and is defined as:

$$RMSE(A,B) = \sqrt{\frac{\sum_{i=1}^{n}(a_i - b_i)^2}{n}} \tag{25}$$

Our example yields $RMSE(A_1, B_1) = \sqrt{1 + 1 + 1 + 1 + 16} = 4.47$ and $RMSE(A_1, B_2) = \sqrt{16 + 4 + 4 + 4 + 4} = 5.66$. Again, we normalize this Root Mean Square Error to get a value between [0, 1].

**Normalized Mean Weighted Order Error (NMWOE).** Next, we calculate the Mean Weighted Order Error (Simske & Vans, 2021) by linearly adjusting the weights depending on how far from the beginning of the gold standard sequence the predicted sequence is. For example, the weight vector for $A_1$ is $W_{A_1} = [5\ 4\ 3\ 2\ 1]$, indicating that the further the position is from the beginning of the sequence, the less significant the actual placement is. The MWOE is thus defined as:

$$MWOE(A,B) = \frac{1}{n}\sum_{i=1}^{n} w_i |a_i - b_i| \tag{26}$$

where $w_i = n - i + 1 \quad \forall w_i \in W$ and $n(W) \doteq n(A)$

As with the first three metrics, the MWOE values are normalized to [0, 1].

**Normalized Clustering Order Error (NCOE).** Finally, we define a metric that acknowledges that documents may form sub-sets within the entire set, but the ordering of those sub-sets may not matter as much as the order within those sub-sets. As a "more forgiving" version of the Levenshtein distance (Levenshtein, 1966), to take this clustering or 'chunking' into consideration, we have:

$$MCOE(A,B) = \frac{1}{n-1}\sum_{i=1}^{n-1} w_i \tag{27}$$

where

$$w_i = \begin{cases} 1 & \text{if } b_i + 1 = b_{i+1} \\ 0 & \text{if } b_i + 1 \neq b_{i+1} \end{cases} \quad \text{if } A = \{1, 2, \ldots, n\}$$

and

$$w_i = \begin{cases} 1 & \text{if } b_i - 1 = b_{i+1} \\ 0 & \text{if } b_i - 1 \neq b_{i+1} \end{cases} \quad \text{if } A = \{n, n-1, \ldots, 1\}$$

Using our sample defined vectors above, we have $MCOE(A_1, B_1) = \frac{1}{4}(1+1+1+0) = 0.75$ and $MCOE(A_1, B_2) = \frac{1}{4}(0+0+1+1) = 0.5$.

**Randomization.** For the above metrics, we are required to have baselines in order to be able to compare certain ordering results. For these purposes, we generated 100 random matrices for the Similarity Matrix Method and 100 random sequences for the Entropy Sequencing Method. The results are in Table 3 and Figure 2. They illustrate the sensitivity of the Normalized Hamming Distance (NHD) to the number of documents to be ordered. The NMHD, NRMSE, and NMWOE metrics, on the other hand, tended to be consistent.

**Table 3. Random Sequences Comparison Metrics for the Similarity Matrix Method and the Entropy Sequencing Method**

| | Similarity Matrix Method | | | | Entropy Sequencing Method | | | |
|---|---|---|---|---|---|---|---|---|
| # Documents | NHD | NMHD | NRMSE | NMWOE | NHD | NMHD | NRMSE | NMWOE |
| 6 | 0.8433 | 0.6489 | 0.6879 | 0.6345 | 0.8433 | 0.6489 | 0.6879 | 0.6345 |
| 15 | 0.9133 | 0.6444 | 0.6810 | 0.6104 | 0.9133 | 0.6444 | 0.6810 | 0.6104 |
| 24 | 0.9492 | 0.6560 | 0.6951 | 0.6180 | 0.9492 | 0.6560 | 0.6951 | 0.6180 |
| 34 | 0.9706 | 0.6581 | 0.6990 | 0.6140 | 0.9706 | 0.6581 | 0.6990 | 0.6140 |
| 44 | 0.9755 | 0.6645 | 0.7029 | 0.6245 | 0.9755 | 0.6645 | 0.7029 | 0.6245 |
| 46 | 0.9804 | 0.6639 | 0.7038 | 0.6226 | 0.9804 | 0.6639 | 0.7038 | 0.6226 |

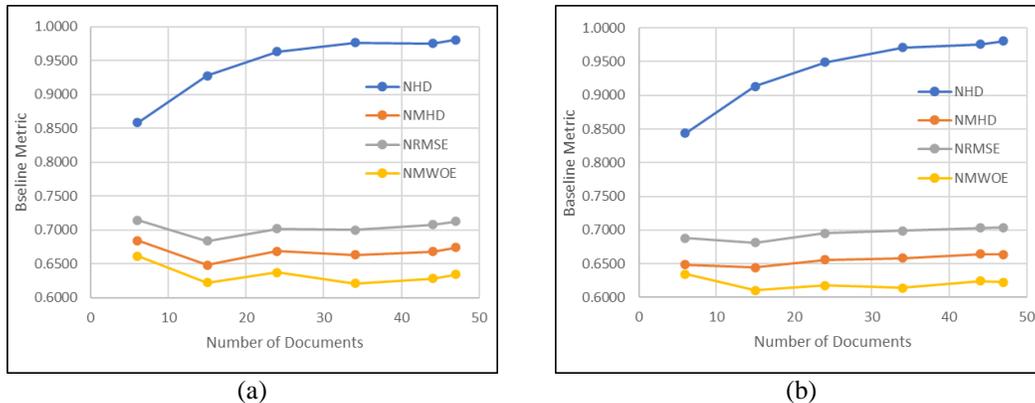

**Figure 2. Graphical Representation of both (a) the Similarity Matrix Method and (b) the Entropy Sequencing Method to Choose a Reasonable Number of Random Matrices and Entropy Sequences**

## 5. Analysis and Results

### 5.1. Assumptions and Initial Parameter Selections

#### 5.1.1. Topics Number Selection: 20% num documents

Methods have been proposed to determine the optimal number of topics fed into LDA algorithms, including (Gan & Qi, 2021), which uses a comprehensive evaluation using perplexity, isolation, stability, and coincidence. However, for simplicity, and based on having a small set of documents (mean < 15) on which to apply LDA, we picked ½ number of documents as number on which to standardize. Some variation in the results of ordering was observed but for our purposes not enough to necessitate a rigorous application of optimality calculations. ½ num documents provided heuristically similar-enough results as other numbers of topics less than the total number of documents. We considered also choosing number of topics to equal the number of documents, but we observed that the Gensim LDA sometimes returned a maximum number of topics fewer than the number of documents.

#### 5.1.2. Summarizer Selection: Luhn

We chose Luhn as the extractive summarizer as it provided the best results in experimentation compared to other Sumy algorithms on the CNN dataset and Wolyn and Simske (Wolyn & Simske, 2022) with the same dataset applying functional analytics. Luhn is based on TF*IDF and relies on word frequency after the removal of stop words and the application of stemming and/or lemmatization. We applied only lemmatization in our case. Other available and considered summarizers include SumBasic, LexRank, Latent Semantic Analysis (LSA), TextRank, Edmundson, and Kullback–Leibler (KL) from the NLTK Sumy library.

### 5.1.3. Summary Percentage Selection: 20%
To realize the usefulness of summarization, it was important to choose a percentage that was not too large as to be too close the being the complete document, and small enough but still be able for the summary to be a good representation of the complete text. For this parameter, we opted to settle on 20%, a reasonable-length summary for a document. Figure 3 shows the effect of various lengths of Luhn summarization on the entropy on four of our test datasets. Each data point corresponds to the difference of entropy calculated for the full document (chapter) and the relative entropy calculated for the summary. A 100% summary indicates an extraction of the complete text and therefore results in zero difference (i.e., identical entropy).

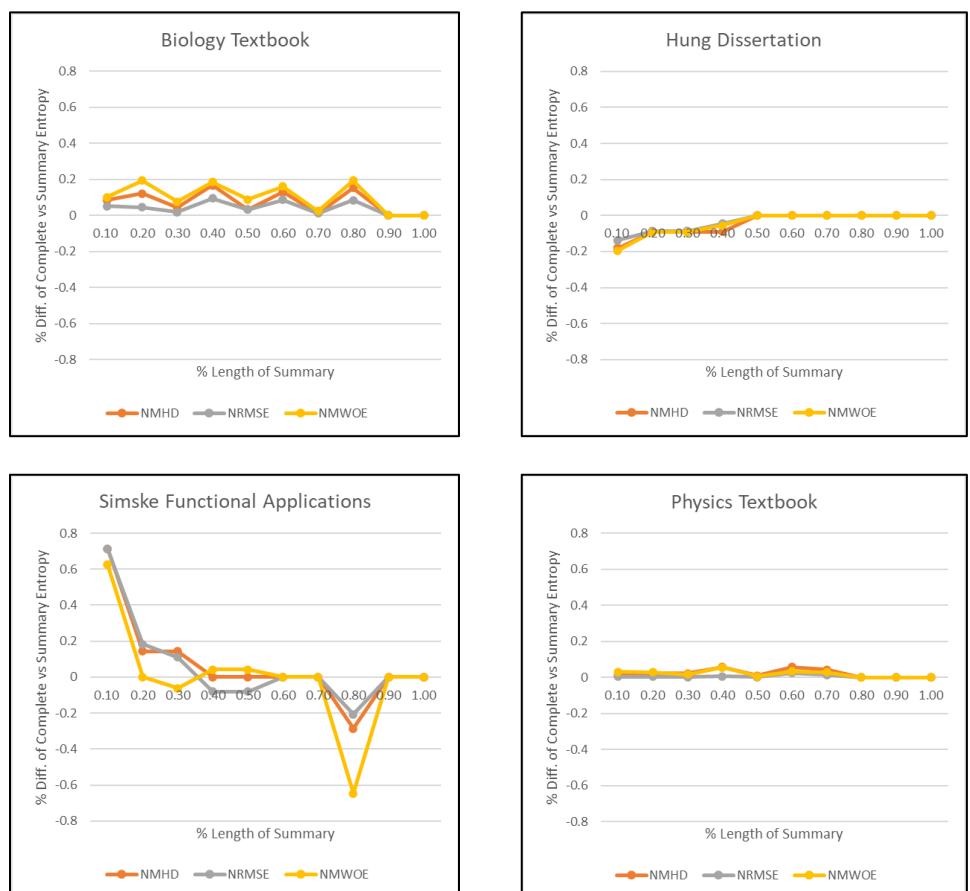

**Figure 3. % Difference of Entropies vs % Length of Summary for Four Datasets Showing that 20% Summarization is a Reasonable Representation**

### 5.2. Document Sequence Ordering
To describe our findings, we first define our corpora based on the two levels. Summarized in Table 4, collections are any of the seven genres of corpora we have under analysis. Each collection is comprised of multiple documents, named depending on the genre. Biographies, novels, dissertations, and textbooks are split into chapters; the Wikipedia collection is comprised of individual articles; courses are divided into lectures; and journals are collections of papers.

## Table 4. Corpus Definitions

| Level 1 Corpus: Collection (Genre) | Level 2 Corpus: Document |
|---|---|
| Biography | Chapter |
| Novel | Chapter |
| Wikipedia | Article |
| Course | Lecture |
| Dissertation | Chapter |
| Journal | Paper |
| Textbook | Chapter |

Table 5 through Table 9 show the statistical p-values obtained using the various metrics applied to each of the ordering schemes described in 4.4( Sequencing Method) applied to the full documents. In each of the metrics, there is a preponderance of the dissertation collections being sequenced by the algorithms much more effectively than random matrices and random entropies. The tables highlight, tiered by the green saturation of the cells, the lower p-values in comparison with random ordering. p-values under our threshold of 0.05 are enclosed in thickened borders for easier identification.

The journal and textbook test collections did not perform as well as dissertations but appear to perform better overall than any of the control collections (biographies, novels, and Wikipedia). The journal collection, in particular, was interesting. We did not anticipate that the order of papers to be given much thought since each paper is typically written by distinct authors and covers a topic that is not dependent on any of the other papers.

Finally, the course collection appeared to not perform any better than the control collections.

Table 10 through Table 14 provide us with similar comparisons as was performed with the complete documents but applied to their 20% Luhn summaries. The results indicate behavior that somewhat mirrored those gotten using the complete documents.

## Table 5. p-values using Normalized Hamming Distance (NHD) Using Complete Documents

| | Sequence Most Similar | | | Sequence Most Similar to Most Recent Document | | | Sequence Least Similar | | | Sequence Most to Least Entropic | | Sequence Least to Most Entropic | |
|---|---|---|---|---|---|---|---|---|---|---|---|---|---|
| | Document Cosine Similarity | Document Jaccard Similarity Coefficient | Document Relative Feature Probability | Document Cosine Similarity | Document Jaccard Similarity Coefficient | Document Relative Feature Probability | Document Cosine Similarity | Document Jaccard Similarity Coefficient | Document Relative Feature Probability | Document Entropy | Document Relative Entropy | Document Entropy | Document Relative Entropy |
| biographies | 0.228356 | 0.084538 | 0.350774 | 0.908754 | 0.497787 | 0.870804 | 0.643875 | 0.286121 | 0.691716 | 0.256162 | 0.769261 | 0.371098 | 0.821014 |
| novels | 0.590040 | 0.738924 | 0.469460 | **0.026721** | 0.080533 | 0.297326 | **0.013106** | 0.466437 | 0.140686 | 0.575800 | 0.635461 | 0.271837 | 0.618457 |
| wikipedia | 0.312741 | 0.889410 | 0.568801 | 0.021324 | 0.264398 | 0.524038 | 0.266696 | 0.760781 | 0.244544 | 0.602492 | 0.206627 | 0.825222 | 0.912171 |
| courses | 0.551021 | 0.917483 | 0.524137 | 0.362253 | 0.657978 | 0.953088 | 0.495507 | 0.853647 | 0.853647 | 0.671785 | 0.147572 | 0.112637 | 0.684796 |
| dissertations | **0.034821** | **0.000137** | 0.575014 | 0.086579 | **0.000363** | 0.863361 | 0.270982 | **0.000000** | 0.117417 | 0.141028 | **0.000000** | 0.256623 | 0.100360 |
| journals | 0.732850 | 0.724387 | 0.743743 | 0.787778 | 0.573654 | 0.681589 | 0.367072 | 0.626801 | 0.456308 | 0.749170 | 0.768452 | 0.378631 | 0.095258 |
| textbooks | 0.246814 | 0.649529 | 0.366325 | 0.490561 | 0.781579 | 0.474684 | 0.554516 | 0.576440 | 0.462811 | 0.808024 | 0.484612 | 0.744095 | 0.981504 |

## Table 6. p-values using Normalized Modified Hamming Distance (NMHD) Using Complete Documents

| | Sequence Most Similar | | | Sequence Most Similar to Most Recent Document | | | Sequence Least Similar | | | Sequence Most to Least Entropic | | Sequence Least to Most Entropic | |
|---|---|---|---|---|---|---|---|---|---|---|---|---|---|
| | Document Cosine Similarity | Document Jaccard Similarity Coefficient | Document Relative Feature Probability | Document Cosine Similarity | Document Jaccard Similarity Coefficient | Document Relative Feature Probability | Document Cosine Similarity | Document Jaccard Similarity Coefficient | Document Relative Feature Probability | Document Entropy | Document Relative Entropy | Document Entropy | Document Relative Entropy |
| biographies | 0.577178 | 0.229963 | 0.527912 | 0.844348 | 0.546910 | 0.609918 | 0.463006 | 0.794145 | 0.542756 | 0.317511 | 0.665922 | 0.813272 | **0.025434** |
| novels | 0.050010 | 0.767241 | 0.129098 | 0.118995 | 0.906349 | 0.555365 | 0.377355 | 0.772027 | 0.439216 | 0.921002 | 0.768506 | 0.821398 | 0.907359 |
| wikipedia | 0.367222 | 0.343003 | 0.351803 | 0.293245 | 0.051232 | **0.037103** | 0.824927 | 0.107483 | 0.844224 | 0.393550 | 0.792485 | 0.790497 | 0.730644 |
| courses | 0.919490 | 0.790422 | 0.262578 | 0.972144 | 0.651563 | 0.305402 | 0.631501 | 0.654018 | 0.468010 | 0.726507 | 0.547774 | 0.581131 | 0.835920 |
| dissertations | 0.069082 | **0.000689** | 0.799823 | 0.067731 | **0.004370** | 0.689430 | 0.457865 | **0.000040** | 0.361081 | **0.000395** | **0.000000** | **0.020272** | **0.000010** |
| journals | 0.253908 | 0.819187 | 0.485483 | 0.053389 | 0.911371 | 0.878143 | 0.324233 | 0.837441 | **0.000454** | **0.013785** | 0.434353 | 0.997746 | **0.049427** |
| textbooks | 0.240665 | 0.539553 | 0.310824 | 0.114284 | 0.462848 | 0.231719 | 0.444043 | 0.176097 | 0.079835 | 0.942386 | 0.587001 | 0.850544 | 0.891166 |

### Table 7. p-values using Normalized Root Mean Square Error (NRMSE) Using Complete Documents

| | Sequence Most Similar | | | Sequence Most Similar to Most Recent Document | | | Sequence Least Similar | | | Sequence Most to Least Entropic | | Sequence Least to Most Entropic | |
|---|---|---|---|---|---|---|---|---|---|---|---|---|---|
| | Document Cosine Similarity | Document Jaccard Similarity Coefficient | Document Relative Feature Probability | Document Cosine Similarity | Document Jaccard Similarity Coefficient | Document Relative Feature Probability | Document Cosine Similarity | Document Jaccard Similarity Coefficient | Document Relative Feature Probability | Document Entropy | Document Relative Entropy | Document Entropy | Document Relative Entropy |
| biographies | 0.629546 | 0.593452 | 0.583256 | 0.657920 | 0.650499 | 0.530071 | 0.401098 | 0.688608 | 0.380862 | 0.084535 | 0.944152 | 0.749576 | 0.431796 |
| novels | **0.022363** | 0.683533 | 0.126024 | 0.171244 | 0.888029 | 0.472315 | 0.164350 | 0.897380 | 0.350093 | 0.895741 | 0.804430 | 0.963095 | 0.823963 |
| wikipedia | 0.217307 | 0.250867 | 0.267704 | 0.165286 | 0.051389 | 0.109846 | 0.420006 | 0.076596 | 0.782539 | 0.753182 | 0.678128 | 0.766324 | 0.659689 |
| courses | 0.916245 | 0.921162 | 0.269017 | 0.857575 | 0.805199 | 0.248095 | 0.702533 | 0.757103 | 0.337917 | 0.796598 | 0.547329 | 0.688651 | 0.639868 |
| dissertations | **0.038183** | 0.079912 | 0.500013 | **0.024958** | 0.114457 | 0.795054 | 0.519246 | **0.002069** | 0.675743 | **0.000026** | **0.000001** | **0.000060** | **0.000008** |
| journals | 0.511914 | 0.735528 | 0.382957 | 0.091737 | 0.855715 | 0.770522 | 0.331624 | 0.989397 | **0.001657** | **0.026938** | 0.172289 | 0.907287 | **0.006459** |
| textbooks | 0.365759 | 0.411945 | 0.192528 | 0.224734 | 0.286839 | 0.290017 | 0.330666 | 0.192804 | 0.098408 | 0.782222 | 0.729438 | 0.812027 | 0.902739 |

### Table 8. p-values using Normalized Mean Weighted Order Error (NMWOE) Using Complete Documents

| | Sequence Most Similar | | | Sequence Most Similar to Most Recent Document | | | Sequence Least Similar | | | Sequence Most to Least Entropic | | Sequence Least to Most Entropic | |
|---|---|---|---|---|---|---|---|---|---|---|---|---|---|
| | Document Cosine Similarity | Document Jaccard Similarity Coefficient | Document Relative Feature Probability | Document Cosine Similarity | Document Jaccard Similarity Coefficient | Document Relative Feature Probability | Document Cosine Similarity | Document Jaccard Similarity Coefficient | Document Relative Feature Probability | Document Entropy | Document Relative Entropy | Document Entropy | Document Relative Entropy |
| biographies | 0.520535 | 0.131520 | 0.358576 | 0.755539 | 0.610948 | 0.553560 | 0.471696 | 0.841336 | 0.399909 | 0.147576 | 0.788441 | 0.976977 | 0.267344 |
| novels | 0.234209 | 0.713077 | 0.153740 | 0.107851 | 0.828441 | 0.707925 | 0.597385 | 0.886057 | 0.667136 | 0.695895 | 0.668951 | 0.852990 | 0.845456 |
| wikipedia | 0.458392 | 0.413507 | 0.069224 | 0.316018 | 0.051647 | 0.081763 | 0.794329 | 0.323296 | 0.902392 | 0.210286 | 0.627129 | 0.575531 | 0.418319 |
| courses | 0.534807 | 0.905641 | 0.489133 | 0.512207 | 0.790435 | 0.524558 | 0.464827 | 0.613111 | 0.300514 | 0.836410 | 0.663377 | 0.739354 | 0.830061 |
| dissertations | **0.002446** | **0.001542** | 0.854322 | **0.003471** | **0.013677** | 0.465288 | **0.014562** | **0.000016** | 0.135995 | **0.045265** | **0.000000** | 0.300952 | **0.000124** |
| journals | 0.449563 | 0.757333 | 0.717591 | 0.067936 | 0.903031 | 0.660533 | 0.501776 | 0.759535 | **0.005140** | **0.010404** | 0.238627 | 0.694174 | **0.008224** |
| textbooks | 0.276074 | 0.518869 | 0.197041 | 0.095879 | 0.416982 | 0.211701 | 0.496457 | 0.212056 | 0.209770 | 0.900966 | 0.802489 | 0.388781 | 0.967295 |

### Table 9. p-values using Normalized Chunking Order Error (NCOE) Using Complete Documents

| | Sequence Most Similar | | | Sequence Most Similar to Most Recent Document | | | Sequence Least Similar | | | Sequence Most to Least Entropic | | Sequence Least to Most Entropic | |
|---|---|---|---|---|---|---|---|---|---|---|---|---|---|
| | Document Cosine Similarity | Document Jaccard Similarity Coefficient | Document Relative Feature Probability | Document Cosine Similarity | Document Jaccard Similarity Coefficient | Document Relative Feature Probability | Document Cosine Similarity | Document Jaccard Similarity Coefficient | Document Relative Feature Probability | Document Entropy | Document Relative Entropy | Document Entropy | Document Relative Entropy |
| biographies | **0.049335** | 0.665037 | **0.006233** | 0.224233 | 0.236676 | 0.112889 | 0.558462 | 0.218909 | 0.578023 | 0.967360 | 0.553383 | 0.533325 | 0.932521 |
| novels | 0.595820 | 0.305372 | 0.647867 | 0.110024 | **0.029919** | **0.043882** | 0.392585 | 0.873744 | 0.757568 | 0.962240 | 0.524080 | 0.280498 | 0.280498 |
| wikipedia | 0.233267 | **0.022088** | **0.000238** | 0.050465 | **0.004672** | 0.200739 | 0.355860 | 0.205555 | **0.018141** | 0.396207 | 0.396125 | 0.218738 | 0.656982 |
| courses | 0.332580 | 0.212584 | **0.004908** | 0.304836 | 0.059078 | 0.173106 | 0.931238 | 0.553257 | **0.001716** | 0.247937 | 0.200896 | 0.453939 | 0.742125 |
| dissertations | **0.001588** | **0.000026** | **0.000835** | **0.000012** | **0.000000** | **0.000000** | **0.019259** | 0.128562 | **0.026373** | **0.000005** | **0.000000** | 0.084246 | 0.253126 |
| journals | 0.609809 | 0.731109 | 0.979748 | 0.365448 | 0.779815 | 0.371775 | 0.587767 | 0.443589 | 0.819172 | 0.759521 | 0.602131 | 0.694422 | 0.123096 |
| textbooks | 0.160006 | 0.987545 | 0.223400 | 0.098754 | 0.247237 | 0.192475 | 0.162837 | 0.492460 | **0.032190** | 0.851619 | 0.627335 | 0.544095 | 0.661581 |

### Table 10. p-values using Normalized Hamming Distance (NHD) Using Summaries

| | Sequence Most Similar | | | Sequence Most Similar to Most Recent Document | | | Sequence Least Similar | | | Sequence Most to Least Entropic | | Sequence Least to Most Entropic | |
|---|---|---|---|---|---|---|---|---|---|---|---|---|---|
| | 0.2 Luhn Summary Cosine Similarity | 0.2 Luhn Summary Jaccard Similarity | 0.2 Luhn Summary Relative Feature | 0.2 Luhn Summary Cosine Similarity | 0.2 Luhn Summary Jaccard Similarity | 0.2 Luhn Summary Relative Feature | 0.2 Luhn Summary Cosine Similarity | 0.2 Luhn Summary Jaccard Similarity | 0.2 Luhn Summary Relative Feature | 0.2 Luhn Summary Entropy | 0.2 Luhn Summary Relative Entropy | 0.2 Luhn Summary Entropy | 0.2 Luhn Summary Relative Entropy |
| biographies | 0.878620 | 0.836961 | 0.988024 | 0.971836 | 0.593015 | 0.885322 | 0.568464 | 0.657124 | 0.894224 | **0.022359** | 0.638684 | 0.970991 | 0.641896 |
| novels | 0.879380 | 0.469460 | 0.415174 | 0.980406 | 0.206233 | 0.925464 | 0.265326 | 0.498021 | 0.631428 | 0.618695 | 0.378003 | 0.075435 | 0.097447 |
| wikipedia | 0.765394 | 0.747738 | 0.984109 | 0.232724 | 0.240344 | 0.977839 | 0.997488 | 0.107976 | 0.686464 | **0.007914** | 0.839332 | 0.302532 | 0.396132 |
| courses | **0.047471** | 0.438847 | 0.524137 | 0.475508 | 0.797012 | 0.735614 | 0.720048 | 0.155822 | 0.801615 | **0.029801** | 0.740475 | 0.845185 | 0.354822 |
| dissertations | 0.639205 | **0.000962** | 0.756733 | **0.010198** | **0.001830** | 0.961145 | 0.387957 | **0.000723** | **0.013289** | 0.155984 | **0.000004** | **0.019682** | **0.005499** |
| journals | 0.674682 | 0.401070 | 0.215598 | 0.469914 | 0.510082 | 0.147423 | 0.798679 | 0.437674 | 0.451458 | 0.582110 | 0.968790 | 0.762069 | 0.977757 |
| textbooks | 0.324939 | 0.687289 | 0.900728 | 0.690002 | 0.818773 | 0.461865 | 0.364087 | 0.501989 | 0.200708 | 0.577508 | 0.732715 | 0.794645 | 0.452393 |

### Table 11. p-values using Normalized Modified Hamming Distance (NMHD) Using Summaries

| | Sequence Most Similar | | | Sequence Most Similar to Most Recent Document | | | Sequence Least Similar | | | Sequence Most to Least Entropic | | Sequence Least to Most Entropic | |
|---|---|---|---|---|---|---|---|---|---|---|---|---|---|
| | 0.2 Luhn Summary Cosine Similarity | 0.2 Luhn Summary Jaccard Similarity | 0.2 Luhn Summary Relative Feature | 0.2 Luhn Summary Cosine Similarity | 0.2 Luhn Summary Jaccard Similarity | 0.2 Luhn Summary Relative Feature | 0.2 Luhn Summary Cosine Similarity | 0.2 Luhn Summary Jaccard Similarity | 0.2 Luhn Summary Relative Feature | 0.2 Luhn Summary Entropy | 0.2 Luhn Summary Relative Entropy | 0.2 Luhn Summary Entropy | 0.2 Luhn Summary Relative Entropy |
| biographies | 0.721221 | 0.586343 | 0.975201 | 0.897959 | **0.010668** | 0.973446 | 0.472357 | 0.885947 | 0.685698 | **0.009736** | 0.639743 | 0.907427 | 0.080054 |
| novels | 0.582587 | 0.683917 | 0.674639 | 0.126923 | 0.118924 | 0.128866 | 0.781686 | 0.200204 | 0.560732 | 0.788122 | 0.775337 | 0.878769 | 0.825104 |
| wikipedia | 0.825073 | 0.057249 | 0.934773 | 0.518708 | 0.078973 | 0.202960 | 0.909645 | 0.448156 | 0.975874 | 0.900361 | 0.555280 | 0.789894 | 0.466545 |
| courses | 0.999153 | 0.626424 | 0.245697 | 0.938251 | 0.543558 | 0.106044 | 0.708308 | 0.894809 | 0.359208 | 0.430468 | 0.662670 | 0.637829 | 0.655176 |
| dissertations | 0.525654 | **0.024735** | 0.484026 | 0.213903 | 0.117246 | 0.233654 | 0.119689 | 0.109529 | 0.066725 | **0.002381** | **0.000017** | 0.354755 | **0.000020** |
| journals | 0.594038 | 0.543224 | 0.448912 | **0.000563** | 0.750873 | 0.997694 | **0.005629** | 0.782818 | **0.018113** | 0.625335 | 0.854827 | 0.392640 | 0.337213 |
| textbooks | 0.133551 | 0.426099 | 0.106180 | 0.179806 | 0.778926 | 0.145418 | 0.912203 | 0.095709 | **0.017267** | 0.881626 | 0.734900 | 0.794727 | 0.369737 |

### Table 12. p-values using Normalized Root Mean Square Error (NRMSE) Using Summaries

| | Sequence Most Similar | | | Sequence Most Similar to Most Recent Document | | | Sequence Least Similar | | | Sequence Most to Least Entropic | | Sequence Least to Most Entropic | |
|---|---|---|---|---|---|---|---|---|---|---|---|---|---|
| | 0.2 Luhn Summary Cosine Similarity | 0.2 Luhn Summary Jaccard Similarity | 0.2 Luhn Summary Relative Feature | 0.2 Luhn Summary Cosine Similarity | 0.2 Luhn Summary Jaccard Similarity | 0.2 Luhn Summary Relative Feature | 0.2 Luhn Summary Cosine Similarity | 0.2 Luhn Summary Jaccard Similarity | 0.2 Luhn Summary Relative Feature | 0.2 Luhn Summary Entropy | 0.2 Luhn Summary Relative Entropy | 0.2 Luhn Summary Entropy | 0.2 Luhn Summary Relative Entropy |
| biographies | 0.785417 | 0.869559 | 0.967650 | 0.861476 | 0.260865 | 0.611780 | 0.467467 | 0.441759 | 0.468163 | 0.051874 | 0.998514 | 0.814924 | 0.189585 |
| novels | 0.575726 | 0.530386 | 0.638932 | 0.155057 | 0.057842 | 0.114418 | 0.977514 | 0.492293 | 0.914711 | 0.653310 | 0.681910 | 0.904659 | 0.848685 |
| wikipedia | 0.803316 | **0.020045** | 0.996358 | 0.366598 | 0.052074 | 0.273325 | 0.810176 | 0.394955 | 0.734786 | 0.825215 | 0.572901 | 0.653748 | 0.449798 |
| courses | 0.751233 | 0.847236 | 0.285134 | 0.684702 | 0.608057 | 0.137415 | 0.714386 | 0.937891 | 0.326870 | 0.695654 | 0.530863 | 0.715642 | 0.543906 |
| dissertations | 0.629176 | 0.388466 | 0.364827 | 0.627675 | 0.517977 | 0.142285 | 0.215263 | 0.329027 | 0.192889 | **0.000036** | **0.000245** | **0.031539** | **0.000156** |
| journals | 0.285996 | 0.764831 | 0.633055 | **0.023548** | 0.750807 | 0.783101 | **0.003648** | 0.874690 | 0.063661 | 0.597804 | 0.712747 | 0.798161 | 0.750693 |
| textbooks | 0.128349 | 0.294633 | 0.064532 | 0.173001 | 0.398213 | 0.121787 | 0.869191 | 0.101932 | **0.031210** | 0.989664 | 0.731015 | 0.855912 | 0.814925 |

### Table 13. p-values using Normalized Mean Weighted Order Error (NMWOE) Using Summaries

| | Sequence Most Similar | | | Sequence Most Similar to Most Recent Document | | | Sequence Least Similar | | | Sequence Most to Least Entropic | | Sequence Least to Most Entropic | |
|---|---|---|---|---|---|---|---|---|---|---|---|---|---|
| | 0.2 Luhn Summary Cosine Similarity | 0.2 Luhn Summary Jaccard Similarity | 0.2 Luhn Summary Relative Feature | 0.2 Luhn Summary Cosine Similarity | 0.2 Luhn Summary Jaccard Similarity | 0.2 Luhn Summary Relative Feature | 0.2 Luhn Summary Cosine Similarity | 0.2 Luhn Summary Jaccard Similarity | 0.2 Luhn Summary Relative Feature | 0.2 Luhn Summary Entropy | 0.2 Luhn Summary Relative Entropy | 0.2 Luhn Summary Entropy | 0.2 Luhn Summary Relative Entropy |
| biographies | 0.901756 | 0.327847 | 0.574296 | 0.961695 | **0.009398** | 0.910288 | 0.509936 | 0.625609 | 0.518414 | **0.003668** | 0.531164 | 0.991973 | 0.139054 |
| novels | 0.937086 | 0.641853 | 0.539971 | 0.354887 | 0.202449 | 0.428176 | 0.958824 | 0.193855 | 0.793131 | 0.531997 | 0.809782 | 0.770972 | 0.736612 |
| wikipedia | 0.873789 | 0.102692 | 0.663987 | 0.974825 | 0.111181 | 0.274073 | 0.715512 | 0.889525 | 0.997698 | 0.795587 | 0.692361 | 0.826526 | 0.604108 |
| courses | 0.839984 | 0.684733 | 0.423926 | 0.780187 | 0.619645 | 0.316626 | 0.459726 | 0.859683 | 0.117283 | 0.568753 | 0.607987 | 0.736786 | 0.662181 |
| dissertations | 0.294352 | **0.040376** | 0.181801 | 0.086314 | 0.202926 | 0.196338 | **0.027014** | **0.015424** | **0.010484** | 0.108029 | **0.000397** | 0.725805 | **0.000416** |
| journals | 0.614625 | 0.417016 | 0.327998 | **0.019466** | 0.826411 | 0.834138 | **0.012418** | 0.871094 | **0.036306** | 0.594913 | 0.435329 | 0.114845 | 0.527728 |
| textbooks | 0.089379 | 0.460573 | 0.088094 | 0.144979 | 0.743817 | 0.078046 | 0.824229 | 0.121796 | **0.039998** | 0.765711 | 0.880805 | 0.358245 | 0.178997 |

### Table 14. p-values using Normalized Chunking Error (NCOE) Using Summaries

| | Sequence Most Similar | | | Sequence Most Similar to Most Recent Document | | | Sequence Least Similar | | | Sequence Most to Least Entropic | | Sequence Least to Most Entropic | |
|---|---|---|---|---|---|---|---|---|---|---|---|---|---|
| | 0.2 Luhn Summary Cosine Similarity | 0.2 Luhn Summary Jaccard Similarity | 0.2 Luhn Summary Relative Feature | 0.2 Luhn Summary Cosine Similarity | 0.2 Luhn Summary Jaccard Similarity | 0.2 Luhn Summary Relative Feature | 0.2 Luhn Summary Cosine Similarity | 0.2 Luhn Summary Jaccard Similarity | 0.2 Luhn Summary Relative Feature | 0.2 Luhn Summary Entropy | 0.2 Luhn Summary Relative Entropy | 0.2 Luhn Summary Entropy | 0.2 Luhn Summary Relative Entropy |
| biographies | 0.159368 | 0.219156 | 0.159368 | 0.220029 | 0.242024 | **0.007630** | 0.578023 | **0.011191** | 0.966244 | 0.289127 | 0.464743 | 0.750348 | 0.828576 |
| novels | 0.561523 | 0.668506 | **0.044186** | 0.151366 | 0.063304 | 0.187955 | 0.204422 | 0.543899 | 0.537511 | 0.315999 | 0.820426 | 0.546991 | 0.960021 |
| wikipedia | **0.034906** | 0.288369 | 0.232869 | **0.000923** | 0.078506 | 0.119946 | 0.089860 | 0.050071 | 0.633398 | 0.313696 | 0.772402 | 0.597734 | 0.255847 |
| courses | 0.371926 | 0.178589 | **0.031249** | 0.141829 | 0.570075 | **0.045856** | 0.156249 | 0.432276 | **0.025880** | 0.480473 | 0.524509 | 0.894819 | 0.562668 |
| dissertations | **0.034149** | **0.000008** | **0.000027** | **0.000700** | **0.000000** | **0.000000** | **0.004225** | 0.153878 | 0.153553 | **0.004777** | **0.000000** | **0.000014** | 0.902559 |
| journals | 0.722225 | 0.668402 | 0.491036 | 0.297695 | 0.770598 | 0.253137 | 0.704238 | 0.992243 | 0.834821 | 0.463024 | 0.327784 | 0.590369 | 0.534577 |
| textbooks | 0.282720 | 0.758413 | 0.238230 | 0.091861 | 0.050987 | **0.009411** | **0.032190** | 0.757809 | 0.597434 | 0.728625 | 0.473357 | 0.664821 | 0.881501 |

To better visualize comparative differences and similarities, a next-level roll-up of the metrics gathered from these tables provides a more concise and clearer comparison. We capture these results by summarizing the results in Table 15 and

Table 16. Here, we observe a strong indication that of all the different collection genres, dissertation document sequences are best predicted by just about any of our five metrics, whether applied to the documents in their entirety (Table 5 - Table 9) or applied to their summaries (Table 10 – Table 14). Table 15 shows the percentage of p-values less than 0.05 and Table 16 shows the mean of the p-values.

### Table 15. Percentage of p-values > 0.05

|  | Full Documents | | | | | Summaries | | | | |
|---|---|---|---|---|---|---|---|---|---|---|
|  | NHD | NMHD | NRMSE | NMWOE | NCOE | NHD | NMHD | NRMSE | NMWOE | NCOE |
| biographies | 1.000 | 0.923 | 1.000 | 1.000 | 0.846 | 0.923 | 0.846 | 1.000 | 0.846 | 0.846 |
| novels | 0.846 | 1.000 | 0.923 | 1.000 | 0.846 | 1.000 | 1.000 | 1.000 | 1.000 | 0.923 |
| wikipedia | 0.923 | 0.923 | 1.000 | 1.000 | 0.692 | 0.923 | 1.000 | 0.923 | 1.000 | 0.846 |
| courses | 1.000 | 1.000 | 1.000 | 1.000 | 0.846 | 0.846 | 1.000 | 1.000 | 1.000 | 0.769 |
| dissertations | 0.615 | 0.462 | 0.462 | 0.308 | 0.231 | 0.385 | 0.692 | 0.692 | 0.538 | 0.231 |
| journals | 1.000 | 0.769 | 0.769 | 0.769 | 1.000 | 1.000 | 0.769 | 0.846 | 0.769 | 1.000 |
| textbooks | 1.000 | 1.000 | 1.000 | 1.000 | 0.923 | 1.000 | 0.923 | 0.923 | 0.923 | 0.846 |

### Table 16. Mean of p-values

|  | Full Documents | | | | | Summaries | | | | |
|---|---|---|---|---|---|---|---|---|---|---|
|  | NHD | NMHD | NRMSE | NMWOE | NCOE | NHD | NMHD | NRMSE | NMWOE | NCOE |
| biographies | 0.522 | 0.535 | 0.563 | 0.525 | 0.434 | 0.734 | 0.604 | 0.599 | 0.539 | 0.377 |
| novels | 0.379 | 0.580 | 0.559 | 0.612 | 0.446 | 0.495 | 0.548 | 0.580 | 0.608 | 0.431 |
| wikipedia | 0.492 | 0.456 | 0.400 | 0.403 | 0.212 | 0.560 | 0.589 | 0.535 | 0.656 | 0.267 |
| courses | 0.599 | 0.642 | 0.653 | 0.631 | 0.324 | 0.513 | 0.601 | 0.598 | 0.591 | 0.340 |
| dissertations | 0.188 | 0.190 | 0.212 | 0.141 | 0.040 | 0.227 | 0.173 | 0.265 | 0.145 | 0.096 |
| journals | 0.591 | 0.466 | 0.445 | 0.444 | 0.605 | 0.569 | 0.489 | 0.542 | 0.433 | 0.588 |
| textbooks | 0.586 | 0.452 | 0.432 | 0.438 | 0.406 | 0.578 | 0.429 | 0.429 | 0.367 | 0.428 |

## 5.3. Effect of Summarization

Another observation is that summaries provide reasonable stand-ins for their complete document counterparts, a verification of the work by Wolyn and Simske (Wolyn & Simske, 2022). We note here that generally, the test collections were better approximated by their summaries for the purposes of ordering compared to the those of the test collections (Table 17).

### Table 17. Difference Between p-values Using Full Documents and Summaries

|  | Summaries | | | | |
|---|---|---|---|---|---|
|  | NHD | NMHD | NRMSE | NMWOE | NCOE |
| biographies | -0.213 | -0.068 | -0.036 | -0.014 | 0.057 |
| novels | -0.117 | 0.031 | -0.022 | 0.005 | 0.015 |
| wikipedia | -0.068 | -0.134 | -0.135 | -0.252 | -0.055 |
| courses | 0.086 | 0.041 | 0.054 | 0.041 | -0.015 |
| dissertations | -0.039 | 0.017 | -0.053 | -0.004 | -0.057 |
| journals | 0.022 | -0.023 | -0.097 | 0.011 | 0.017 |
| textbooks | 0.009 | 0.023 | 0.004 | 0.071 | -0.022 |

Furthermore, close inspection shows a tight correlation of results as indicated by our whisker-plots in Figure 4. Consolidated distributions of differences in p-values in Figure 5 indicate tightness that approximates a normal distribution in the case of Figure 5-(c) and Figure 5-(d). Figure 5-(a), Figure 5-(b), and Figure 5-(e) approximate the upper half of a sinc function.

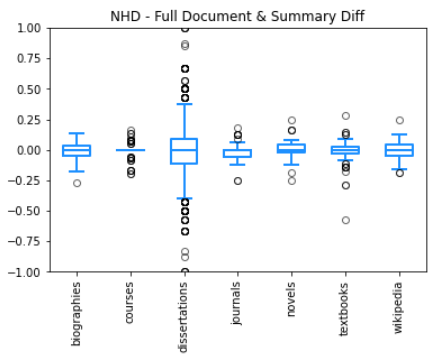

(a)

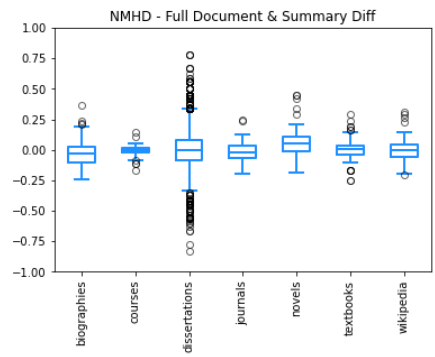

(b)

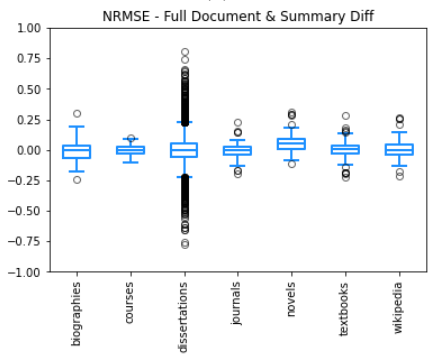

(c)

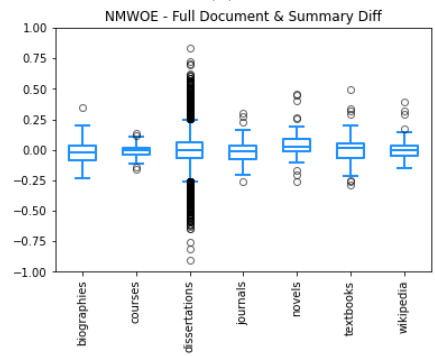

(d)

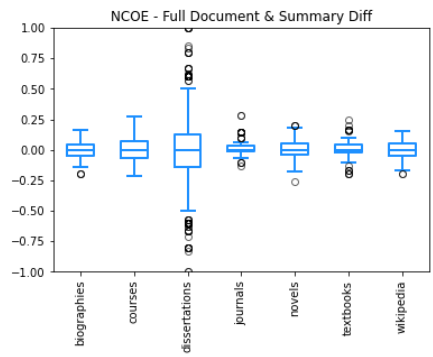

(e)

**Figure 4. Distribution of Differences Between Complete Document p-values and Summary p-values by Collection**

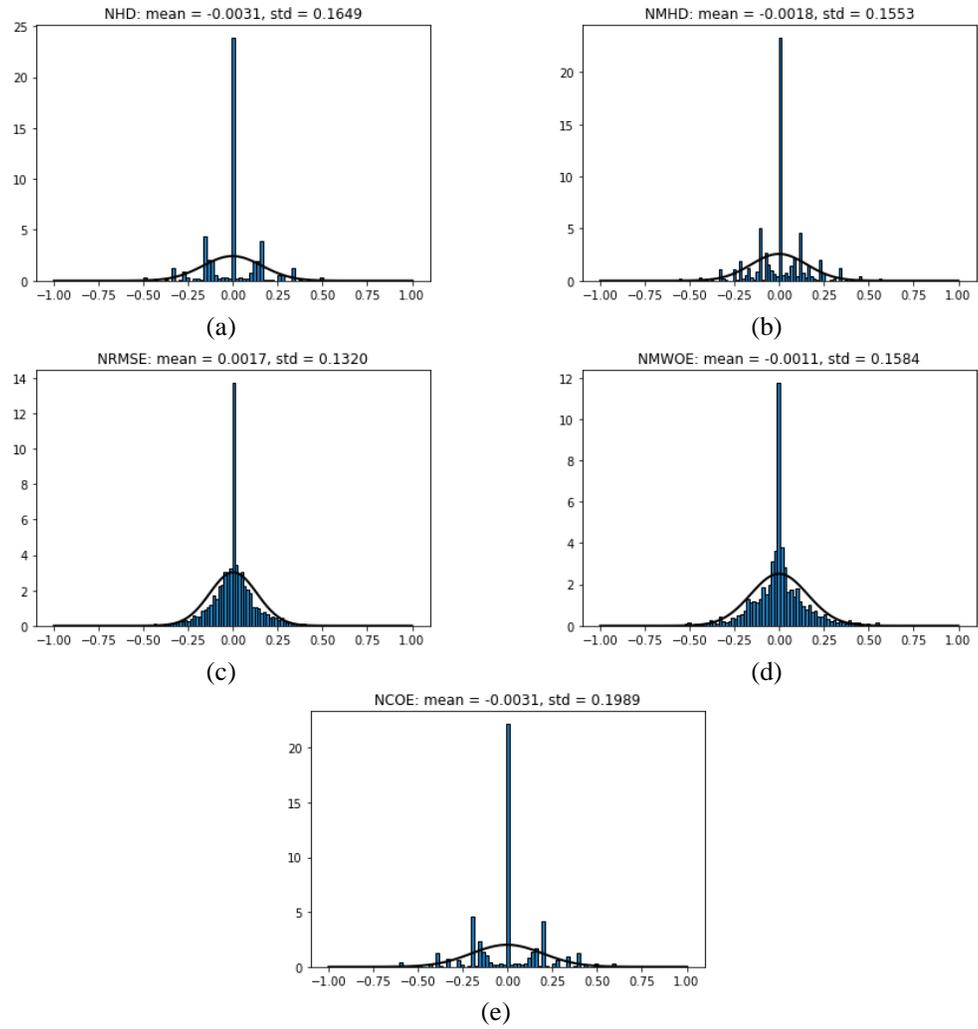

**Figure 5. Distribution of Consolidated p-values of Full Documents minus p-values of Summary Documents**

## 5.4. Analysis of Various Departmental Dissertations

Compared to the other genres, because of their availability and accessibility from Colorado State University, we have collected more dissertations than all of the others combined. But because many departments are represented (we have 32 that had five or more dissertations that had 6 or more chapters), we were also interested in seeing if predictability varied across them. To quantify this, we used ANOVA tests for all the sequencing schemes (including for both complete documents and summaries) we investigated measured with all our metrics and determined that indeed there were variations. Rather than showing all the results, we present a sample set of results for the metric NMWOE for *Sequence Most to Least Entropic* using *Document Relative Entropy* with only the pairs in which the p-value ≤ 0.5.

**Table 18. Sample ANOVA Test Results: NMWOE, Sequence Most to Least Entropic, Document Relative Entropy; p-value: 0.314339**

| Department 1 | Department 2 | p-value |
| --- | --- | --- |
| Agricultural Biology and Economics | Atmospheric Science | 0.010100 |
| Agricultural Biology and Economics | Biochemistry _ Molecular Biology | 0.016900 |
| Agricultural Biology and Economics | Environmental and Radiological Health Sciences | 0.047600 |
| Agricultural Biology and Economics | Human Dimensions of Natural Resources | 0.046200 |
| Atmospheric Science | Computer Science | 0.004200 |
| Atmospheric Science | Economics | 0.019500 |
| Atmospheric Science | Food Science and Human Nutrition | 0.047800 |
| Atmospheric Science | Mathematics | 0.024000 |
| Atmospheric Science | Statistics | 0.014200 |
| Biochemistry _ Molecular Biology | Computer Science | 0.006700 |
| Biochemistry _ Molecular Biology | Economics | 0.029800 |
| Biochemistry _ Molecular Biology | Mathematics | 0.034500 |
| Biochemistry _ Molecular Biology | Statistics | 0.022300 |
| Civil and Environmental Engineering | Computer Science | 0.019000 |
| Civil and Environmental Engineering | Statistics | 0.044400 |
| Civil and Environmental Engineering | Systems Engineering | 0.048500 |
| Computer Science | Environmental and Radiological Health Sciences | 0.021800 |
| Computer Science | Forest and Rangeland Stewardship | 0.021900 |
| Computer Science | Human Dimensions of Natural Resources | 0.016200 |
| Computer Science | Physics | 0.045400 |
| Computer Science | Political Science | 0.031100 |
| Economics | Human Dimensions of Natural Resources | 0.048500 |
| Human Dimensions of Natural Resources | Statistics | 0.037400 |

Using the Bonferroni Correction (Bonferroni, 1936) and calculating the t-statistic, the p-values are sorted and summarized in Table 19. For this particular set of results, we note that Computer Science dissertations are most different from the rest of the departments.

**Table 19. Sorted t-static p-values after a Bonferroni Correction based on Table 18**

| Department | Mean p-value Compared to Other Departments |
|---|---|
| Computer Science | 0.235975 |
| Civil and Environmental Engineering | 0.250743 |
| Atmospheric Science | 0.267396 |
| Human Dimensions of Natural Resources | 0.278377 |
| Biochemistry _ Molecular Biology | 0.295775 |
| Agricultural Biology and Economics | 0.296260 |
| Forest and Rangeland Stewardship | 0.384956 |
| Environmental and Radiological Health Sciences | 0.394341 |
| Political Science | 0.408809 |
| Statistics | 0.432523 |
| Mathematics | 0.456586 |
| Economics | 0.458738 |
| Physics | 0.472531 |
| Biomedical Sciences | 0.514174 |
| Systems Engineering | 0.535665 |
| Chemistry | 0.550397 |
| Mechanical Engineering | 0.568558 |
| Food Science and Human Nutrition | 0.571663 |
| Journalism _ Media Communication | 0.579678 |
| Geosciences | 0.581272 |
| Electrical and Computer Engineering | 0.586312 |
| Biology | 0.590579 |
| Soil _ Crop Sciences | 0.599033 |
| Animal Sciences | 0.606219 |
| Psychology | 0.610589 |
| Chemical and Biological Engineering | 0.614384 |
| Clinical Sciences | 0.615356 |
| Horticulture _ Landscape Architecture | 0.617040 |
| Health and Exercise Science | 0.621763 |
| Sociology | 0.629826 |
| Microbiology, Immunology, and Pathology | 0.637283 |
| Fish, Wildlife, and Conservation Biology | 0.656991 |

### 5.5. Limitations

In **Sequence Most to Least KL-Divergent**, we made an assumption that there was topic balance, and therefore used a uniform distribution $q$ for comparison. It is quite possible that the topic distributions followed a normal, Zipf, or some other distribution.

# 6. Conclusions

To avoid subjective interpretation of results, we identified and used five different objective metrics, comparing results using our algorithms with those results using random sequencing, using p-values obtained from both t-test and ANOVA statistical tests.

### 6.1. Predicting Existing Order

We have demonstrated through this research, employing an extensive comparison of results from using educational material and control document sets and using multiple metrics, that we are able to, better than randomly (and in some instances much better than randomly) predict the existing order of these documents (i.e., the order intended by the author or editor). As expected, predicting the order of control documents within the sets (biographies, novels, and randomly picked Wikipedia articles on a broad subject) was not successful using our methods, but predicting the orders of the test documents (courses, dissertations, journals, and textbooks) were, especially with dissertations.

Perhaps of most surprise to the author, journal article order could be somewhat predicted, indicating that journal editors do not simply publish articles in an issue in random order. The results of our experiments on textbook chapters can be thought of as mirroring what textbook ordering should be. Though an explicit ordering of chapters from Chapter 1 through Chapter n is often implied, chapters can be consumed in random fashion. The author, after all, does not remember a time when he read a textbook from cover to cover. Perhaps the same can be said of a university course, which oftentimes mirrors the chapters in a textbook.

### 6.2. Curriculum Development and Proposing Order for Comprehension

As an extension to Section 6.1, if order wasn't correctly predicted, we submit that it is feasible that more suitable orders exist, and therefore our algorithms could be used to automatically generate proper sequences. For example, we would use the **Sequence Most to Least Entropic** algorithm as a way to order reading or present a curriculum based on a textbook or series of textbooks by acknowledging that more generic materials (i.e., material with more concepts covered) are prerequisites to more in-depth topics. For example, an introductory book on quantum computing may cover qubits, a brief history of quantum mechanics, post-quantum cryptography, quantum supremacy, and Cuda, while a more advanced book may cover simply the mathematics of post-quantum cryptography, but in depth.

### 6.3. Understanding the Results of Summarization

Though not unexpected, particularly in light of extensive work by Simske (Simske & Vans, 2021) on summarization and text analytics which have demonstrated the effectiveness of functional analytics, it is of note that our work in document ordering is also applicable using summarized documents, at least for the fixed hyper-parameters that we have chosen in Section 5.1 where we settled on the number of topics at 20% the number of documents, using Luhn as the extractive summarizer of choice, and limiting the size of the summary at 20% of the length (number of sentences) of the full document.

# 7. Further Experimentation and Research

Our research, while extensive, provides a set of approaches for additional research. While certain hyper-parametric assumptions were made with reasonable justifications, they are not substitutes for extensive research and experimentation on what the ideal parameters are. The following are some areas in which additional work is warranted:

Using different summarizers. We used Luhn as a summarizer. However, because our research worked on news articles (an entirely different genre of text from the ones for this section), using Luhn may not have been optimal. In future work, additional extractive summarizers and even abstractive summarizers may be incorporated for comparison. In addition, a meta-algorithmic summarizer such as the one we developed using Tessellation and Recombination with Expert Decisioner may provide superior results when combining several extractive and/or abstractive summarizers.

Optimizing number of topics. There are well-known methods of picking the optimal number of topics for K-means clustering. While we settled on a number that was 20% the size of the number of documents, the ideal number would depend on the text itself and would likely be different depending on the genre and contents of the text.

Optimizing summary percentage. Here, we reasoned that summaries that were about 20% of the full text were reasonable. The ideal percentage may be different, and likely is dependent on the type of documents summarized.

Transcripted lectures. Instead of just the lecture slides, adding the transcript will provide an additional investigation worth pursuing. When we originally conceptualized using course lecture slides, we desired to include the professor's transcribed lecture as part of the lecture slides. Unfortunately, those were not readily available, and we settled for just the lecture slides.

Ordering song tracks. While CDs are now mostly relegated to the Smithsonian, artists still do release most of their songs as parts of "albums." It would be interesting to see if it would be feasible to order song tracks in an album using lyrics and/or chords. Artists can be particular about these sequences (Ruiz, 2021).

Using the Predictive Selection meta-algorithm for Section 5.4. Since we noticed differences among department dissertations, we could use Predictive Selection to "pre-classify" the dissertations by department and run the best (for that department) reading order approach to detect if the overall system reading order accuracy improves.

Large Language Models (LLMs). With the prevalence of the use of ChatGPT and other LLMs, it would be prudent in future to experiment with functional comparisons of results from these models.

# Funding

This research received no external funding.

# Declaration of Conflicting Interest

The authors declare no conflict of interest.

# References

Al-Muhaideb, S., & Menai, M. E. B. (2011). Evolutionary computation approaches to the Curriculum Sequencing problem. *Natural Computing*, *10*(2), 891–920. https://doi.org/10.1007/s11047-010-9246-5

Arani, A. A. K., Karami, H., Gharehpetian, G. B., & Hejazi, M. S. A. (2017). Review of Flywheel Energy Storage Systems structures and applications in power systems and microgrids. *Renewable and Sustainable Energy Reviews*, *69*, 9–18. https://doi.org/10.1016/j.rser.2016.11.166

Blei, D. M., Ng, A. Y., & Jordan, M. I. (2003). Latent dirichlet allocation. *The Journal of Machine Learning Research*, *3*(null), 993–1022.

Bonferroni, C. E. (1936). *Teoria statistica delle classi e calcolo delle probabilità*. Pubblicazioni del R Istituto Superiore di Scienze Economiche e Commerciali di Firenze.

*Cosine Similarity—An overview*. (n.d.). ScienceDirect. Retrieved June 14, 2023, from https://www.sciencedirect.com/topics/computer-science/cosine-similarity

Forestier, S., Portelas, R., Mollard, Y., & Oudeyer, P.-Y. (2022). Intrinsically motivated goal exploration processes with automatic curriculum learning. *The Journal of Machine Learning Research*, *23*(1), 152:6818-152:6858.

Gan, J., & Qi, Y. (2021). Selection of the Optimal Number of Topics for LDA Topic Model-Taking Patent Policy Analysis as an Example. *Entropy (Basel, Switzerland)*, *23*(10), 1301. https://doi.org/10.3390/e23101301

Hamming, R. (2012). Classical Error-Correcting Codes: Machines should work. People should think. In D. C. Marinescu & G. M. Marinescu (Eds.), *Classical and Quantum Information* (pp. 345–454). Academic Press. https://doi.org/10.1016/B978-0-12-383874-2.00004-7

Koutrika, G., Liu, L., & Simske, S. (2015). Generating reading orders over document collections. *2015 IEEE 31st International Conference on Data Engineering*, 507–518. https://doi.org/10.1109/ICDE.2015.7113310

Kurt, W. (2019). Bayesian statistics the fun way: Understanding statistics and probability with Star Wars, LEGO, and Rubber Ducks. No Starch Press.

Kurt, W. (2017, May 10). *Kullback-Leibler Divergence Explained*. Count Bayesie. http://www.countbayesie.com/blog/2017/5/9/kullback-leibler-divergence-explained

Levenshtein, V. I. (1966). Binary Codes Capable of Correcting Deletions, Insertions and Reversals. *Soviet Physics Doklady*, *10*, 707–710. https://ui.adsabs.harvard.edu/abs/1966SPhD...10..707L

Margunayasa, I. G., Dantes, N., Marhaeni, A. A. I. N., & Suastra, I. W. (2019). The Effect of Guided Inquiry Learning and Cognitive Style on Science Learning Achievement. *International Journal of Instruction*, *12*(1), 737–750. https://eric.ed.gov/?id=EJ1201135

Menai, M. E. B., Alhunitah, H., & Al-Salman, H. (2018). Swarm intelligence to solve the curriculum sequencing problem. *Computer Applications in Engineering Education*, *26*(5), 1393–1404. https://doi.org/10.1002/cae.22046

Niwattanakul, S., Singthongchai, J., Naenudorn, E., & Wanapu, S. (2013). Using of Jaccard Coefficient for Keywords Similarity. *Proceedings of the International MultiConference of Engineers and Computer Scientists*. Proceedings of the International MultiConference of Engineers and Computer Scientists, Hong Kong.

Nurnberger-Haag, J., & Thompson, C. A. (2023). Simplest Shapes First! But Let's Use Cognitive Science to Reconceive and Specify What "Simple" Means. *Mind, Brain, and Education*, *17*(1), 5–19. https://doi.org/10.1111/mbe.12338

Pedregosa, F., Varoquaux, G., Gramfort, A., Michel, V., Thirion, B., Grisel, O., Blondel, M., Prettenhofer, P., Weiss, R., Dubourg, V., Vanderplas, J., Passos, A., & Cournapeau, D. (2011). Scikit-learn: Machine Learning in Python. *Journal of Machine Learning Research*, *12*, 2825–2830.

Pennington, J., Socher, R., & Manning, C. (2014). GloVe: Global Vectors for Word Representation. *Proceedings of the 2014 Conference on Empirical Methods in Natural Language Processing (EMNLP)*, 1532–1543. https://doi.org/10.3115/v1/D14-1162

*Project Gutenberg*. (n.d.). Project Gutenberg. Retrieved October 1, 2022, from https://www.gutenberg.org/

Python Software Foundation. (2022, October). *Sumy 0.11.0*. The Python Package Index (PyPI). https://pypi.org/project/sumy/

Radim Řehůřek. (n.d.). *Word2Vec Model*. Gensim: Topic Modelling for Humans. Retrieved November 1, 2022, from https://radimrehurek.com/gensim/auto_examples/tutorials/run_word2vec.html

Řehůřek, R., & Sojka, P. (2010). Software Framework for Topic Modelling with Large Corpora. *Proceedings of the LREC 2010 Workshop on New Challenges for NLP Frameworks*, 45–50.

Richards, J. C. (2021). *Curriculum Development in Language Teaching* (2nd ed.). Cambridge University Press. https://doi.org/10.1017/9781009024556

Ruiz, J. V. (2021, December 17). Grammy winner explains why Adele is right – album tracks should not be shuffled. *Big Think*. https://bigthink.com/high-culture/dont-shuffle-albums/

*Scipy.stats.entropy 1.9.1*. (n.d.). The SciPy Community. Retrieved September 1, 2022, from https://docs.scipy.org/doc/scipy/reference/generated/scipy.stats.entropy.html

Shannon, C. E. (1948). A Mathematical Theory of Communication. *Bell System Technical Journal*, *27*(3), 379–423. https://doi.org/10.1002/j.1538-7305.1948.tb01338.x

Simske, S., & Vans, M. (2021). *Functional Applications of Text Analytics Systems*. River Publishers.

Thinniyam, R. (2014). *On Statistical Sequencing of Document Collections* [PhD thesis, University of Toronto]. https://tspace.library.utoronto.ca/bitstream/1807/68101/1/Thinniyam_Ramya_201406_PhD_thesis.pdf

Verma, P., Pal, S., & Om, H. (2019). A Comparative Analysis on Hindi and English Extractive Text Summarization. *ACM Transactions on Asian and Low-Resource Language Information Processing*, *18*(3), 1–39. https://doi.org/10.1145/3308754

Villanueva, A. N., Jr. (2023). *Optimizing text analytics and document automation with meta-algorithmic systems engineering* [D.Eng. dissertation, Colorado State University]. ProQuest. https://www.proquest.com/docview/2853933036

Wolyn, S., & Simske, S. J. (2022). Summarization assessment methodology for multiple corpora using queries and classification for functional evaluation. Integrated Computer-Aided Engineering, 29(3), 227–239. https://doi.org/10.3233/ICA-220680